\documentclass{article}
\pdfoutput=1

\usepackage{PRIMEarxiv}

\usepackage[utf8]{inputenc} % allow utf-8 input
\usepackage[T1]{fontenc}    % use 8-bit T1 fonts
\usepackage{hyperref}       % hyperlinks
\usepackage{url}            % simple URL typesetting
\usepackage{booktabs}       % professional-quality tables
\usepackage{amsfonts}       % blackboard math symbols
\usepackage{nicefrac}       % compact symbols for 1/2, etc.
\usepackage{microtype}      % microtypography
\usepackage{lipsum}
\usepackage{fancyhdr}       % header
\usepackage{graphicx}       % graphics
\graphicspath{{media/}}     % organize your images and other figures under media/ folder
\usepackage[hang,flushmargin]{footmisc}

\usepackage{graphicx}
\usepackage[style=numeric]{biblatex} %Imports biblatex package
\title{The Landscape of Emerging AI Agent Architectures for Reasoning, Planning, and Tool Calling: A Survey
}

\author{
  Tula Masterman* \\
  Neudesic, an IBM Company \\
  \texttt{tula.masterman@neudesic.com} \\
   \And
  Sandi Besen* \\
  IBM \\
  \texttt{sandi.besen@ibm.com} \\
  \AND
  Mason Sawtell* \\
  Neudesic, an IBM Company \\
  \texttt{mason.sawtell@neudesic.com} \\
  \\\relax
  * Denotes Equal Contribution
  \And
  Alex Chao \\
  Microsoft \\
  \texttt{achao@microsoft.com} \\
}

\begin{document}
\maketitle

\let\thefootnote\relax\footnotetext{The opinions expressed in this paper are solely those of the authors and do not necessarily reflect the views or policies of their respective employers.}

\begin{abstract}
This survey paper examines the recent advancements in AI agent
implementations, with a focus on their ability to achieve complex goals
that require enhanced reasoning, planning, and tool execution
capabilities. The primary objectives of this work are to a) communicate
the current capabilities and limitations of existing AI agent
implementations, b) share insights gained from our
observations of these systems in action, and c) suggest important
considerations for future developments in AI agent design. We achieve
this by providing overviews of single-agent and multi-agent
architectures, identifying key patterns and divergences in design
choices, and evaluating their overall impact on accomplishing a provided
goal. Our contribution outlines key themes when selecting an agentic
architecture, the impact of leadership on agent systems, agent
communication styles, and key phases for planning, execution, and
reflection that enable robust AI agent systems.
\end{abstract}

% keywords can be removed

\section{Introduction}

Since the launch of ChatGPT, many of the first wave of generative AI applications have been a variation of a chat over a corpus of documents using the Retrieval Augmented Generation (RAG) pattern. While there is a lot of activity in making RAG systems more robust, various groups are starting to build what the next generation of AI applications will look like, centralizing on a common theme: agents.

Beginning with investigations into recent foundation models like GPT-4 and popularized through open-source projects like AutoGPT and BabyAGI, the research community has experimented with building autonomous agent-based systems \cite{nakajima_yoheinakajimababyagi_2024,birr_autogptp_2024}.

As opposed to zero-shot prompting of a large language model where a user types into an open-ended text field and gets a result without additional input, agents allow for more complex interaction and orchestration. In particular, agentic systems have a notion of planning, loops, reflection and other control structures that heavily leverage the model's inherent reasoning capabilities to accomplish a task end-to-end. Paired with the ability to use tools, plugins, and function calling, agents are empowered to do more general-purpose work.

Among the community, there is a current debate on whether single or multi-agent systems are best suited for solving complex tasks. While single agent architectures excel when problems are well-defined and feedback from other agent-personas or the user is not needed, multi-agent architectures tend to thrive more when collaboration and multiple distinct execution paths are required.

\

\begin{figure}[h]
    \centering
    \includegraphics[width=1\linewidth]{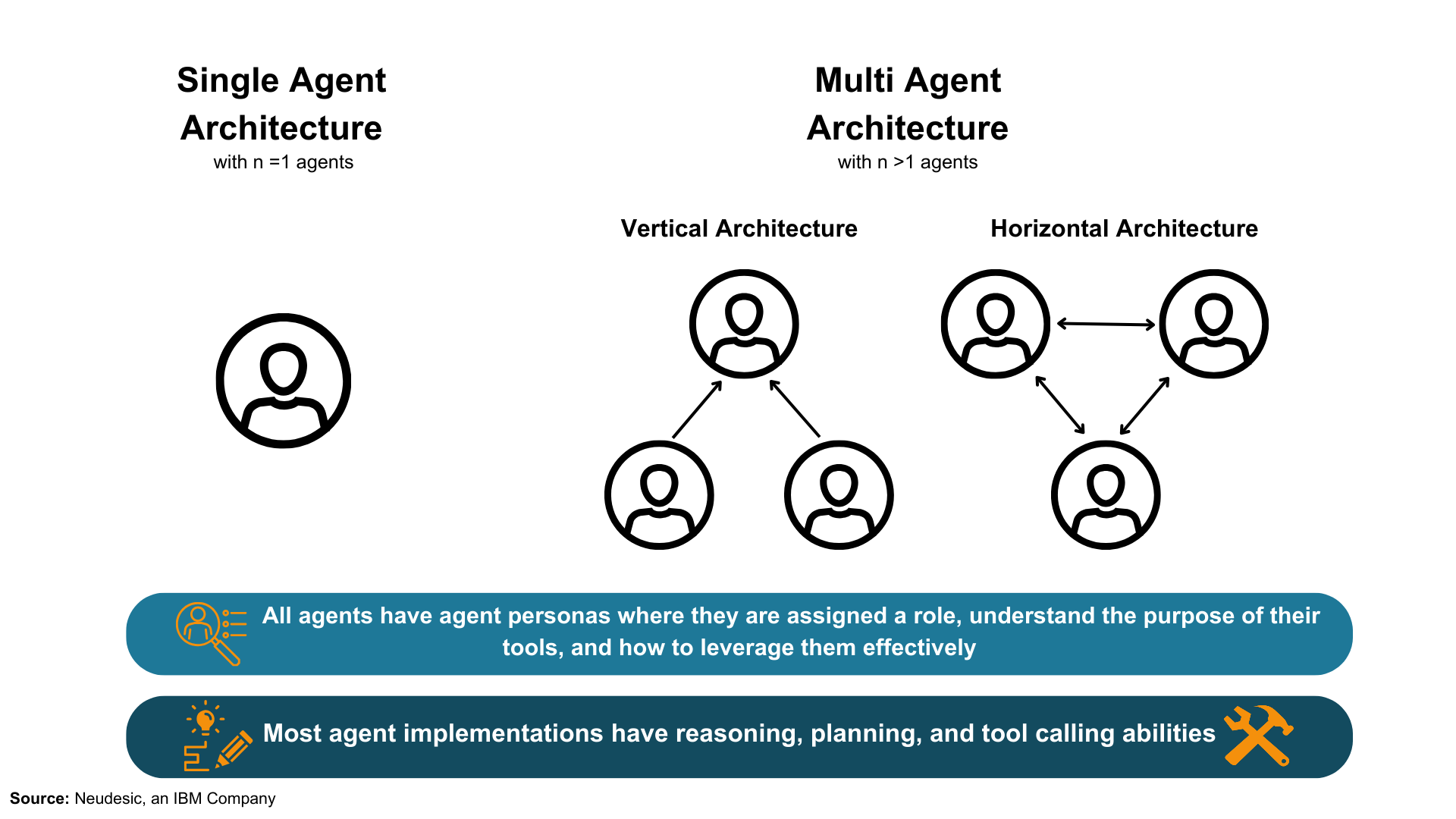}
    \caption{A visualization of single and multi-agent architectures with their underlying features and abilities}
    \label{fig:enter-label}
\end{figure}

\subsection{Taxonomy}

\textbf{Agents}.
AI agents are language model-powered entities able to plan and take actions to execute goals over multiple iterations. AI agent architectures are either comprised of a single agent or multiple agents working together to solve a problem.

Typically, each agent is given a persona and access to a variety of tools that will help them accomplish their job either independently or as part of a team. Some agents also contain a memory component, where they can save and load information outside of their messages and prompts. In this paper, we follow the definition of agent that consists of ``brain, perception, and action'' \cite{xi2023rise}. These components satisfy the minimum requirements for agents to understand, reason, and act on the environment around them.

\textbf{Agent Persona}.
An agent persona describes the role or personality that the agent should take on, including any other instructions specific to that agent. Personas also contain descriptions of any tools the agent has access to. They make the agent aware of their role, the purpose of their tools, and how to leverage them effectively. Researchers have found that ``shaped personality verifiably influences Large Language Model (LLM) behavior in common downstream (i.e. subsequent) tasks, such as writing social media posts'' \cite{serapiogarcía2023personality}. Solutions that use multiple agent personas to solve problems also show significant improvements compared to Chain-of-Thought (CoT) prompting where the model is asked to break down its plans step by step \cite{wang2024unleashing,wei_chain--thought_2023}.

\textbf{Tools}. In the context of AI agents, tools represent any functions that the model can call. They allow the agent to interact with external data sources by pulling or pushing information to that source. An example of an agent persona and associated tools is a professional contract writer. The writer is given a persona explaining their role and the types of tasks it must accomplish. It is also given tools related to adding notes to a document, reading an existing document, or sending an email with a final draft.

\textbf{Single Agent Architectures}.
These architectures are powered by one language model and will perform all the reasoning, planning, and tool execution on their own. The agent is given a system prompt and any tools required to complete their task. In single agent patterns there is no feedback mechanism from other AI agents; however, there may be options for humans to provide feedback that guides the agent.

\textbf{Multi-Agent Architectures}.
These architectures involve two or more agents, where each agent can utilize the same language model or a set of different language models. The agents may have access to the same tools or different tools. Each agent typically has their own persona.

Multi-agent architectures can have a wide variety of organizations at any level of complexity. In this paper, we divide them into two primary categories: vertical and horizontal. It is important to keep in mind that these categories represent two ends of a spectrum, where most existing architectures fall somewhere between these two extremes.

\textbf{Vertical Architectures}.
In this structure, one agent acts as a leader and has other agents report directly to them. Depending on the architecture, reporting agents may communicate exclusively with the lead agent. Alternatively, a leader may be defined with a shared conversation between all agents. The defining features of vertical architectures include having a lead agent and a clear division of labor between the collaborating agents.

\textbf{Horizontal Architectures}. In this structure, all the agents are treated as equals and are part of one group discussion about the task. Communication between agents occurs in a shared thread where each agent can see all messages from the others. Agents also can volunteer to complete certain tasks or call tools, meaning they do not need to be assigned by a leading agent. Horizontal architectures are generally used for tasks where collaboration, feedback and group discussion are key to the overall success of the task \cite{chen_agentverse_2023}.

\section{Key Considerations for Effective Agents}

\subsection{Overview}
Agents are designed to extend language model capabilities to solve real-world challenges. Successful implementations require robust problem-solving capabilities enabling agents to perform well on novel tasks. To solve real-world problems effectively, agents require the ability to reason and plan as well as call tools that interact with an external environment. In this section we explore why reasoning, planning, and tool calling are critical to agent success.

\subsection{The Importance of Reasoning and Planning}
Reasoning is a fundamental building block of human cognition, enabling people to make decisions, solve problems, and understand the world around us. AI agents need a strong ability to reason if they are to effectively interact with complex environments, make autonomous decisions, and assist humans in a wide range of tasks. This tight synergy between ``acting'' and ``reasoning'' allows new tasks to be learned quickly and enables robust decision making or reasoning, even under previously unseen circumstances or information uncertainties \cite{yao_react_2023}. Additionally, agents need reasoning to adjust their plans based on new feedback or information learned.

If agents lacking reasoning skills are tasked with acting on straightforward tasks, they may misinterpret the query, generate a response based on a literal understanding, or fail to consider multi-step implications. 

Planning, which requires strong reasoning abilities, commonly falls into one of five major approaches: task decomposition, multi-plan selection, external module-aided planning, reflection and refinement and memory-augmented planning \cite{huang2024understanding}. These approaches allow the model to either break the task down into sub tasks, select one plan from many generated options, leverage a preexisting external plan, revise previous plans based on new information, or leverage external information to improve the plan.

Most agent patterns have a dedicated planning step which invokes one or more of these techniques to create a plan before any actions are executed. For example, Plan Like a Graph (PLaG) is an approach that represents plans as directed graphs, with multiple steps being executed in parallel \cite{lin_graph-enhanced_2024,yao_tree_2023}. This can provide a significant performance increase over other methods on tasks that contain many independent subtasks that benefit from asynchronous execution.

\subsection{The Importance of Effective Tool Calling}
One key benefit of the agent abstraction over prompting base language models is the agents' ability to solve complex problems by calling multiple tools. These tools enable the agent to interact with external data sources, send or retrieve information from existing APIs, and more. Problems that require extensive tool calling often go hand in hand with those that require complex reasoning.

Both single-agent and multi-agent architectures can be used to solve challenging tasks by employing reasoning and tool calling steps. Many methods use multiple iterations of reasoning, memory, and reflection to effectively and accurately complete problems \cite{liu_llm_2024,shinn_reflexion_2023,yao_react_2023}. They often do this by breaking a larger problem into smaller subproblems, and then solving each one with the appropriate tools in sequence.

Other works focused on advancing agent patterns highlight that while breaking a larger problem into smaller subproblems can be effective at solving complex tasks, single agent patterns often struggle to complete the long sequence required \cite{shi_learning_2024,gao_efficient_2024}.

Multi-agent patterns can address the issues of parallel tasks and robustness since individual agents can work on individual subproblems. Many multi-agent patterns start by taking a complex problem and breaking it down into several smaller tasks. Then, each agent works independently on solving each task using their own independent set of tools.

\section{Single Agent Architectures}

\subsection{Overview}
In this section, we highlight some notable single agent methods  such as ReAct, RAISE, Reflexion, AutoGPT + P, and LATS. Each of these methods contain a dedicated stage for reasoning about the problem before any action is taken to advance the goal. We selected these methods based on their contributions to the reasoning and tool calling capabilities of agents.

\subsection{Key Themes}
We find that successful goal execution by agents is contingent upon proper planning and self-correction \cite{yao_react_2023,liu_llm_2024,shinn_reflexion_2023,birr_autogptp_2024}. Without the ability to self-evaluate and create effective plans, single agents may get stuck in an endless execution loop and never accomplish a given task or return a result that does not meet user expectations \cite{yao_react_2023}. We find that single agent architectures are especially useful when the task requires straightforward function calling and does not need feedback from another agent \cite{shi_learning_2024}.

\subsection{Examples}

\textbf{ReAct.}
In the ReAct (Reason + Act) method, an agent first writes a thought about the given task. It then performs an action based on that thought, and the output is observed. This cycle can repeat until the task is complete \cite{yao_react_2023}. When applied to a diverse set of language and decision-making tasks, the ReAct method demonstrates improved effectiveness compared to zero-shot prompting on the same tasks. It also provides improved human interoperability and trustworthiness because the entire thought process of the model is recorded. When evaluated on the HotpotQA dataset, the ReAct method only hallucinated 6\% of the time, compared to 14\% using the chain of thought (CoT) method \cite{wei_chain--thought_2023,yao_react_2023}.

However, the ReAct method is not without its limitations. While intertwining reasoning, observation, and action improves trustworthiness, the model can repetitively generate the same thoughts and actions and fail to create new thoughts to provoke finishing the task and exiting the ReAct loop. Incorporating human feedback during the execution of the task would likely increase its effectiveness and applicability in real-world scenarios.

\begin{figure}[h]
    \centering
    \includegraphics[width=6in,height=2.5374in]{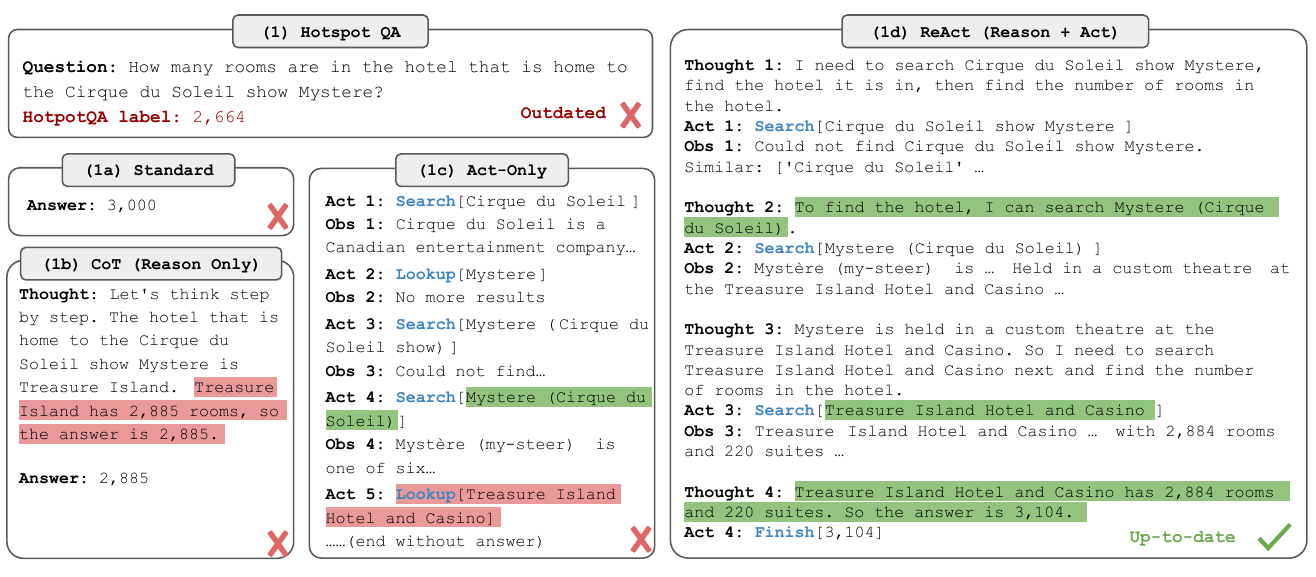}
    \caption{An example of the ReAct method compared to other methods \cite{yao_react_2023}}
    \label{fig:ReAct}
\end{figure}

\textbf{RAISE.}
The RAISE method is built upon the ReAct method, with the addition of a memory mechanism that mirrors human short-term and long-term memory \cite{liu_llm_2024}. It does this by using a scratchpad for short-term storage and a dataset of similar previous examples for long-term storage.

By adding these components, RAISE improves upon the agent's ability to retain context in longer conversations. The paper also highlights how fine-tuning the model results in the best performance for their task, even when using a smaller model. They also showed that RAISE outperforms ReAct in both efficiency and output quality.

While RAISE significantly improves upon existing methods in some respects, the researchers also highlighted several issues. First, RAISE struggles to understand complex logic, limiting its usefulness in many scenarios. Additionally, RAISE agents often hallucinated with respect to their roles or knowledge. For example, a sales agent without a clearly defined role might retain the ability to code in Python, which may enable them to start writing Python code instead of focusing on their sales tasks. These agents might also give the user misleading or incorrect information. This problem was addressed by fine-tuning the model, but the researchers still highlighted hallucination as a limitation in the RAISE implementation.

\begin{figure}[ht]
    \centering
    \includegraphics[width=6in,height=3.4125in, trim = 0cm 7cm 0cm 1cm]{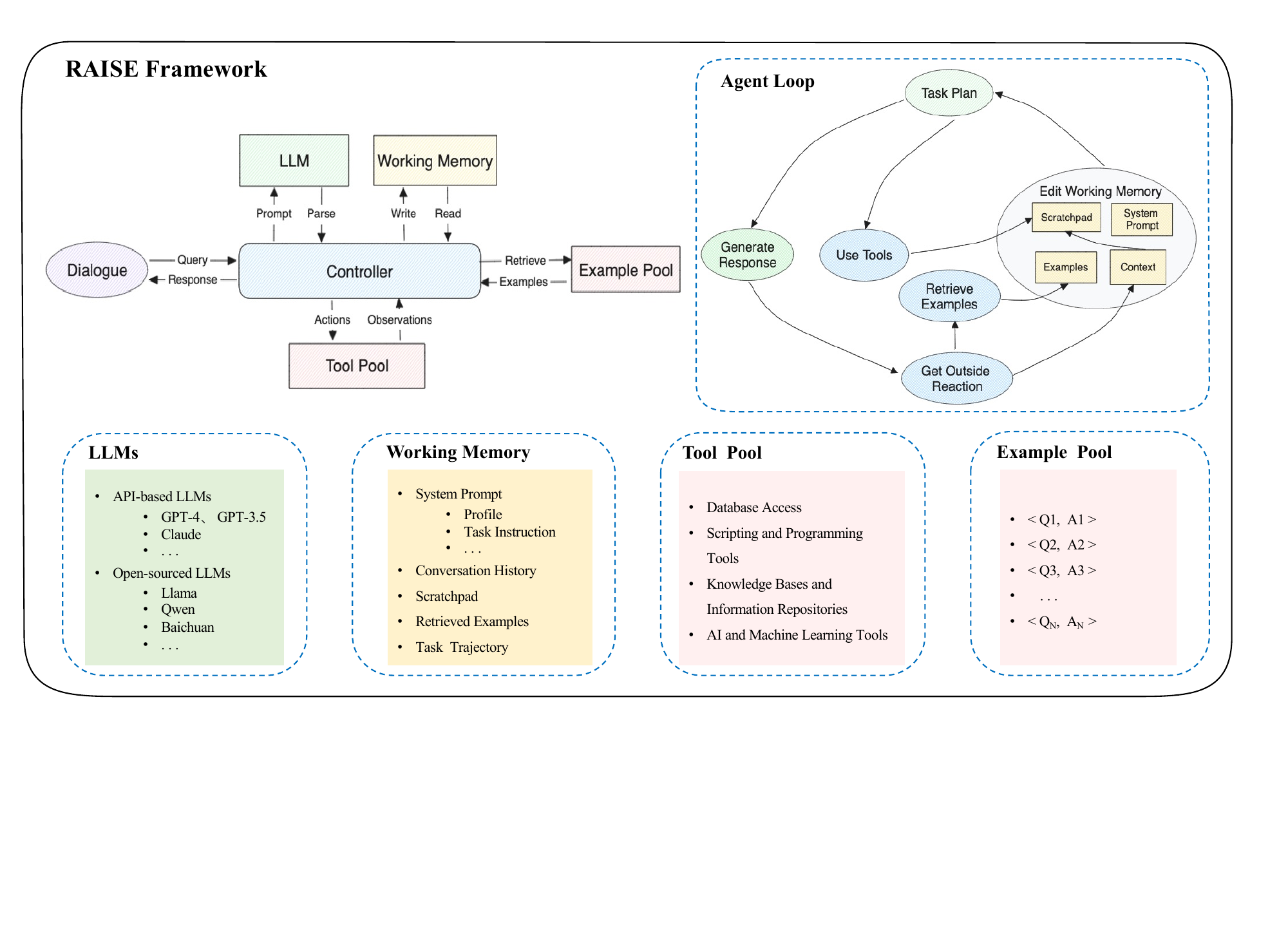}
    \caption{A diagram showing the RAISE method \cite{liu_llm_2024}}
    \label{fig:RAISE}
\end{figure}

\textbf{Reflexion.}
Reflexion is a single-agent pattern that uses self-reflection through linguistic feedback \cite{shinn_reflexion_2023}. By utilizing metrics such as success state, current trajectory, and persistent memory, this method uses an LLM evaluator to provide specific and relevant feedback to the agent. This results in an improved success rate as well as reduced hallucination compared to Chain-of-Thought and ReAct.

Despite these advancements, the Reflexion authors identify various limitations of the pattern. Primarily, Reflexion is susceptible to ``non-optimal local minima solutions''. It also uses a sliding window for long-term memory, rather than a database. This means that the volume of long-term memory is limited by the token limit of the language model. Finally, the researchers identify that while Reflexion surpasses other single-agent patterns, there are still opportunities to improve performance on tasks that require a significant amount of diversity, exploration, and reasoning.

\textbf{AUTOGPT + P.}
AutoGPT + P (Planning) is a method that addresses reasoning limitations for agents that command robots in natural language \cite{birr_autogptp_2024}. AutoGPT+P combines object detection and Object Affordance Mapping (OAM) with a planning system driven by a LLM. This allows the agent to explore the environment for missing objects, propose alternatives, or ask the user for assistance with reaching its goal.

AutoGPT+P starts by using an image of a scene to detect the objects present. A language model then uses those objects to select which tool to use, from four options: Plan Tool, Partial Plan Tool, Suggest Alternative Tool, and Explore Tool. These tools allow the robot to not only generate a full plan to complete the goal, but also to explore the environment, make assumptions, and create partial plans.

However, the language model does not generate the plan entirely on its own. Instead, it generates goals and steps to work aside a classical planner which executes the plan using Planning Domain Definition Language (PDDL). The paper found that ``LLMs currently lack the ability to directly translate a natural language instruction into a plan for executing robotic tasks, primarily due to their constrained reasoning capabilities'' \cite{birr_autogptp_2024}. By combining the LLM planning capabilities with a classical planner, their approach significantly improves upon other purely language model-based approaches to robotic planning.

As with most first of their kind approaches, AutoGPT+P is not without its drawbacks. Accuracy of tool selection varies, with certain tools being called inappropriately or getting stuck in loops. In scenarios where exploration is required, the tool selection sometimes leads to illogical exploration decisions like looking for objects in the wrong place. The framework also is limited in terms of human interaction, with the agent being unable to seek clarification and the user being unable to modify or terminate the plan during execution.

\begin{figure}[h]
    \centering
    \includegraphics[width=5in,height=1.6145in]{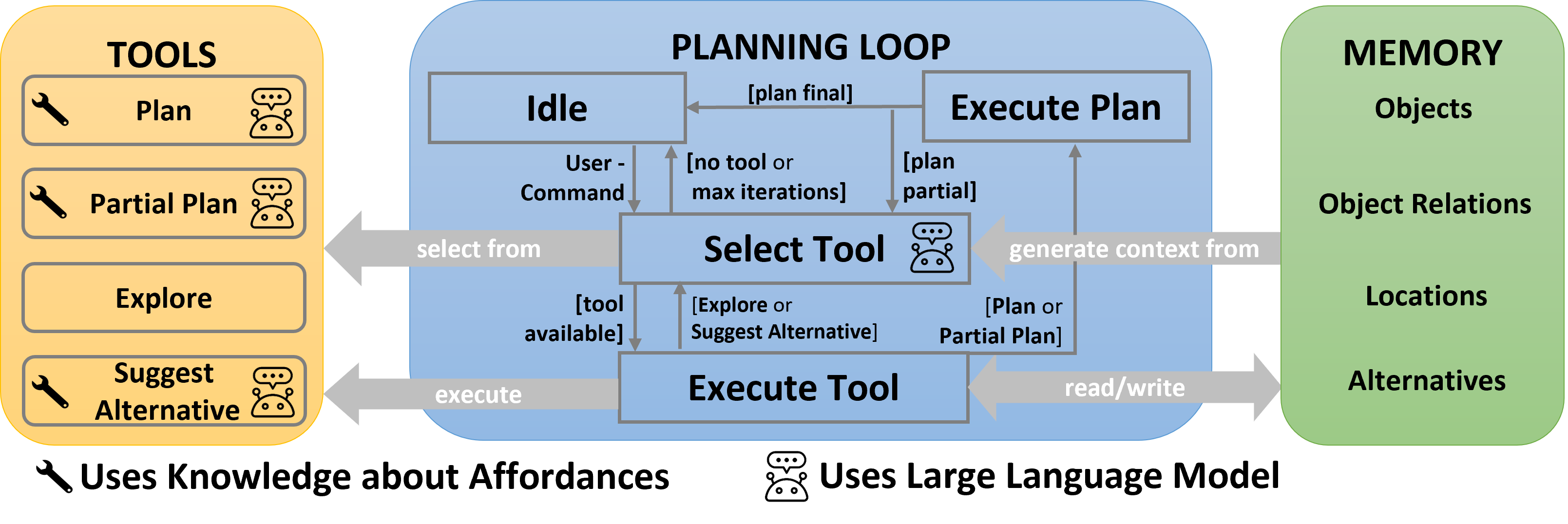}
    \caption{A diagram of the AutoGPT+P method \cite{birr_autogptp_2024}}
    \label{fig:AutoGPT+P}
\end{figure}

\textbf{LATS.}
Language Agent Tree Search (LATS) is a single-agent method that synergizes planning, acting, and reasoning by using trees \cite{zhou_language_2023}. This technique, inspired by Monte Carlo Tree Search, represents a state as a node and taking an action as traversing between nodes. It uses LM-based heuristics to search for possible options, then selects an action using a state evaluator.

When compared to other tree-based methods, LATS implements a self-reflection reasoning step that dramatically improves performance. When an action is taken, both environmental feedback as well as feedback from a language model is used to determine if there are any errors in reasoning and propose alternatives. This ability to self-reflect combined with a powerful search algorithm makes LATS perform extremely well on various tasks.

However, due to the complexity of the algorithm and the reflection steps involved, LATS often uses more computational resources and takes more time to complete than other single-agent methods \cite{zhou_language_2023}. The paper also uses relatively simple question answering benchmarks and has not been tested on more robust scenarios that involve involving tool calling or complex reasoning.

\section{Multi Agent Architectures}

\subsection{Overview}
In this section, we examine a few key studies and sample frameworks with multi-agent architectures, such as Embodied LLM Agents Learn to Cooperate in Organized Teams, DyLAN, AgentVerse, and MetaGPT. We highlight how these implementations facilitate goal execution through inter-agent communication and collaborative plan execution. This is not intended to be an exhaustive list of all agent frameworks, our goal is  to provide broad coverage of key themes and examples related to multi-agent patterns.

\subsection{Key Themes}
Multi-agent architectures create an opportunity for both the intelligent division of labor based on skill and helpful feedback from a variety of agent personas. Many multi-agent architectures work in stages where teams of agents are created and reorganized dynamically for each planning, execution, and evaluation phase \cite{chen_agentverse_2023,guo2024embodied,liu2023dynamic}. This reorganization provides superior results because specialized agents are employed for certain tasks, and removed when they are no longer needed. By matching agents roles and skills to the task at hand, agent teams can achieve greater accuracy and decrease time to meet the goal. Key features of effective multi-agent architectures include clear leadership in agent teams, dynamic team construction, and effective information sharing between team members so that important information does not get lost in superfluous chatter.

\subsection{Examples}
\textbf{Embodied LLM Agents Learn to Cooperate in Organized Teams.}
Research by Guo et al. demonstrates the impact of a lead agent on the overall effectiveness of the agent team \cite{guo2024embodied}. This architecture contains a vertical component through the leader agent, as well as a horizontal component from the ability for agents to converse with other agents besides the leader. The results of their study demonstrate that agent teams with an organized leader complete their tasks nearly 10\% faster than teams without a leader.

Furthermore, they discovered that in teams without a designated leader, agents spent most of their time giving orders to one another (\textasciitilde50\% of communication), splitting their remaining time between sharing information, or requesting guidance. Conversely, in teams with a designated leader, 60\% of the leader's communication involved giving directions, prompting other members to focus more on exchanging and requesting information. Their results demonstrate that agent teams are most effective when the leader is a human.

\begin{figure}[h]
    \centering
    \includegraphics[width=6in]{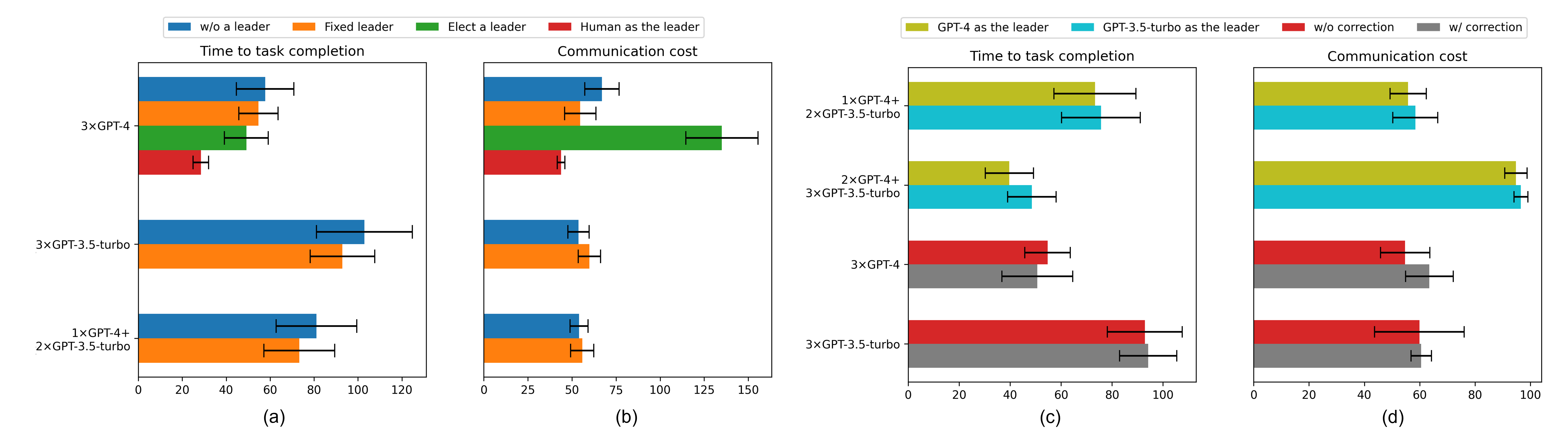}
    \caption{Agent teams with a designated leader achieve superior performance \cite{guo2024embodied}}.
    \label{fig:org_results_13}
\end{figure}

Beyond team structure, the paper emphasizes the importance of employing a ``criticize-reflect'' step for generating plans, evaluating performance, providing feedback, and re-organizing the team \cite{guo2024embodied}. Their results indicate that agents with a dynamic team structure with rotating leadership provide the best results, with both the lowest time to task completion and the lowest communication cost on average. Ultimately, leadership and dynamic team structures improve the overall team's ability to reason, plan, and perform tasks effectively.

\textbf{DyLAN.}
The Dynamic LLM-Agent Network (DyLAN) framework creates a dynamic agent structure that focuses on complex tasks like reasoning and code generation \cite{liu2023dynamic}. DyLAN has a specific step for determining how much each agent has contributed in the last round of work and only moves top contributors the next round of execution. This method is horizontal in nature since agents can share information with each other and there is no defined leader. DyLAN shows improved performance on a variety of benchmarks which measure arithmetic and general reasoning capabilities. This highlights the impact of dynamic teams and demonstrates that by consistently re-evaluating and ranking agent contributions, we can create agent teams that are better suited to complete a given task.

\textbf{AgentVerse.}
Multi-agent architectures like AgentVerse demonstrate how distinct phases for group planning can improve an AI agent's reasoning and problem-solving capabilities \cite{chen_agentverse_2023}. AgentVerse contains four primary stages for task execution: recruitment, collaborative decision making, independent action execution, and evaluation. This can be repeated until the overall goal is achieved. By strictly defining each phase, AgentVerse helps guide the set of agents to reason, discuss, and execute more effectively.

As an example, the recruitment step allows agents to be removed or added based on the progress towards the goal. This helps ensure that the right agents are participating at any given stage of problem solving. The researchers found that horizontal teams are generally best suited for collaborative tasks like consulting, while vertical teams are better suited for tasks that require clearer isolation of responsibilities for tool calling.

\begin{figure}[h]
    \centering
    \includegraphics[width=6in]{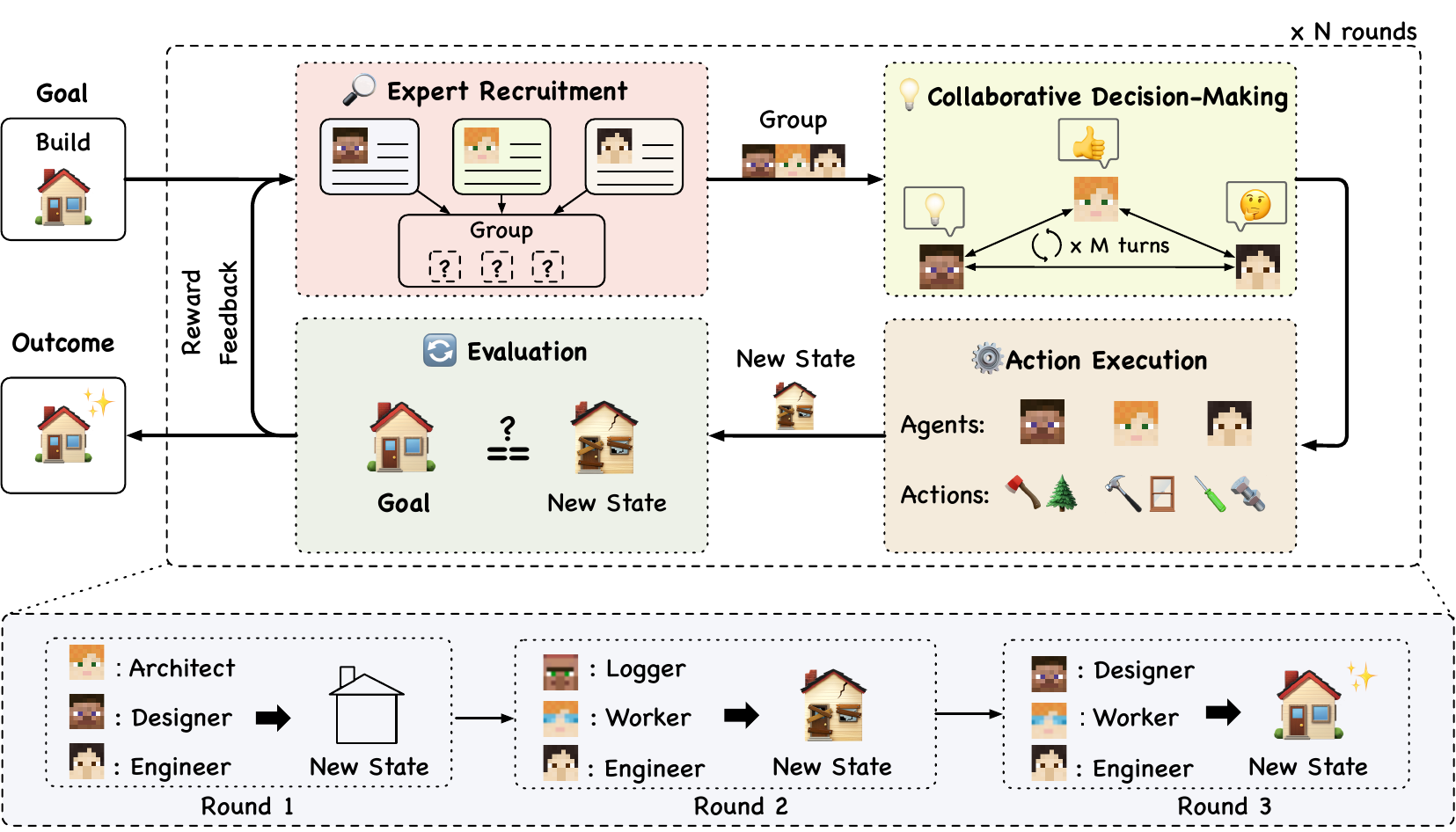}
    \caption{A diagram of the AgentVerse method \cite{chen_agentverse_2023}}
    \label{fig:agentverse}
\end{figure}

\textbf{MetaGPT.}
Many multi-agent architectures allow agents to converse with one another while collaborating on a common problem. This conversational capability can lead to chatter between the agents that is superfluous and does not further the team goal. MetaGPT addresses the issue of unproductive chatter amongst agents by requiring agents to generate structured outputs like documents and diagrams instead of sharing unstructured chat messages \cite{hong2023metagpt}.

Additionally, MetaGPT implements a ''publish-subscribe'' mechanism for information sharing. This allows all the agents to share information in one place, but only read information relevant to their individual goals and tasks. This streamlines the overall goal execution and reduces conversational noise between agents. When compared to single-agent architectures on the HumanEval and MBPP benchmarks, MetaGPT's multi-agent architecture demonstrates significantly better results.

\section{Discussion and Observations}

\subsection{Overview}
In this section we discuss the key themes and impacts of the design choices exhibited in the previously outlined agent patterns. These patterns serve as key examples of the growing body of research and implementation of AI agent architectures. Both single and multi-agent architectures seek to enhance the capabilities of language models by giving them the ability to execute goals on behalf of or alongside a human user. Most observed  agent implementations broadly follow the plan, act, and evaluate process to iteratively solve problems.

We find that both single and multi-agent architectures demonstrate compelling performance on complex goal execution. We also find that across architectures clear feedback, task decomposition, iterative refinement, and role definition yield improved agent performance.

\subsection{Key Findings}

\textbf{Typical Conditions for Selecting a Single vs Multi-Agent
Architecture.}
Based on the aforementioned agent patterns, we find that single-agent patterns are generally best suited for tasks with a narrowly defined list of tools and where processes are well-defined. Single agents are also typically easier to implement since only one agent and set of tools needs to be defined. Additionally, single agent architectures do not face limitations like poor feedback from other agents or distracting and unrelated chatter from other team members. However, they may get stuck in an execution loop and fail to make progress towards their goal if their reasoning and refinement capabilities are not robust.

Multi-agent architectures are generally well-suited for tasks where feedback from multiple personas is beneficial in accomplishing the task. For example, document generation may benefit from a multi-agent architecture where one agent provides clear feedback to another on a written section of the document. Multi-agent systems are also useful when parallelization across distinct tasks or workflows is required. Crucially, Wang et. al finds that multi-agent patterns perform better than single agents in scenarios when no examples are provided \cite{wang_rethinking_2024}. By nature, multi-agent systems are more complex and often benefit from robust conversation management and clear leadership.

While single and multi-agent patterns have diverging capabilities in terms of scope, research finds that ``multi-agent discussion does not necessarily enhance reasoning when the prompt provided to an agent is sufficiently robust'' \cite{wang_rethinking_2024}. This suggests that those implementing agent architectures should decide between single or multiple agents based on the broader context of their use case, and not based on the reasoning capabilities required.

\textbf{Agents and Asynchronous Task Execution.}
While a single agent can initiate multiple asynchronous calls simultaneously, its operational model does not inherently support the division of responsibilities across different execution threads. This means that, although tasks are handled asynchronously, they are not truly parallel in the sense of being autonomously managed by separate decision-making entities. Instead, the single agent must sequentially plan and execute tasks, waiting for one batch of asynchronous operations to complete before it can evaluate and move on to the next step. Conversely, in multi-agent architectures, each agent can operate independently, allowing for a more dynamic division of labor. This structure not only facilitates simultaneous task execution across different domains or objectives but also allows individual agents to proceed with their next steps without being hindered by the state of tasks handled by others, embodying a more flexible and parallel approach to task management.

\textbf{Impact of Feedback and Human Oversight on Agent Systems.} 
When solving a complex problem, it is extremely unlikely that one provides a correct, robust solution on their first try. Instead, one might pose a potential solution before criticizing it and refining it. One could also consult with someone else and receive feedback from another perspective. The same idea of iterative feedback and refinement is essential for helping agents solve complex problems.

This is partially because language models tend to commit to an answer earlier in their response, which can cause a `snowball effect' of increasing diversion from their goal state \cite{zhang_how_2023} . By implementing feedback, agents are much more likely to correct their course and reach their goal.

Additionally, the inclusion of human oversight improves the immediate outcome by aligning the agent's responses more closely with human expectations, mitigating the potential for agents to delve down an inefficient or invalid approach to solving a task. As of today, including human validation and feedback in the agent architecture yields more reliable and trustworthy results \cite{feng2024large,guo2024embodied}.

Language models also exhibit sycophantic behavior, where they ``tend to mirror the user's stance, even if it means forgoing the presentation of an impartial or balanced viewpoint'' \cite{park_ai_2023}. Specifically, the AgentVerse paper describes how agents are susceptible to feedback from other agents, even if the feedback is not sound. This can lead the agent team to generate a faulty plan which diverts them from their objective  \cite{chen_agentverse_2023}. Robust prompting can help mitigate this, but those developing agent applications should be aware of the risks when implementing user or agent feedback systems.

\textbf{Challenges with Group Conversations and Information Sharing.}
One challenge with multi-agent architectures lies in their ability to
intelligently share messages between agents. Multi-agent patterns have a greater tendency to get caught up in niceties and ask one another things like ``how are you'', while single agent patterns tend to stay focused on the task at hand since there is no team dynamic to manage. The extraneous dialogue in multi-agent systems can impair both the agent's ability to reason effectively and execute the right tools, ultimately distracting the agents from the task and decreasing team efficiency. This is especially true in a horizontal architecture, where agents typically share a group chat and are privy to every agent's message in a conversation. Message subscribing or filtering improves multi-agent performance by ensuring agents only receive information relevant to their tasks.

In vertical architectures, tasks tend to be clearly divided by agent skill which helps reduce distractions in the team. However, challenges arise when the leading agent fails to send critical information to their supporting agents and does not realize the other agents aren't privy to necessary information. This failure can lead to confusion in the team or hallucination in the results. One approach to address this issue is to explicitly include information about access rights in the system prompt so that the agents have contextually appropriate interactions.

\textbf{Impact of Role Definition and Dynamic Teams.}
Clear role definition is critical for both single and multi-agent architectures. In single-agent architectures role definition ensures that the agent stays focused on the provided task, executes the proper tools, and minimizes hallucination of other capabilities. Similarly, role definition in multi-agent architectures ensures each agent knows what it's responsible for in the overall team and does not take on tasks outside of their described capabilities or scope. Beyond individual role definition, establishing a clear group leader also improves the overall performance of multi-agent teams by streamlining task assignment. Furthermore, defining a clear system prompt for each agent can minimize excess chatter by prompting the agents not to engage in unproductive communication.

Dynamic teams where agents are brought in and out of the system based on need have also been shown to be effective. This ensures that all agents participating in the planning or execution of tasks are fit for that round of work.

\subsection{Summary}
Both single and multi-agent patterns exhibit strong performance on a variety of complex tasks involving reasoning and tool execution. Single agent patterns perform well when given a defined persona and set of tools, opportunities for human feedback, and the ability to work iteratively towards their goal. When constructing an agent team that needs to collaborate on complex goals, it is beneficial to deploy agents with at least one of these key elements: clear leader(s), a defined planning phase and opportunities to refine the plan as new information is learned, intelligent message filtering, and dynamic teams whose agents possess specific skills relevant to the current sub-task. If an agent architecture employs at least one of these approaches it is likely to result in increased performance compared to a single agent architecture or a multi-agent architecture without these tactics.

\section{Limitations of Current Research and Considerations for Future Research}

\subsection{Overview}
In this section we examine some of the limitations of agent research today and identify potential areas for improving AI agent systems. While agent architectures have significantly enhanced the capability of language models in many ways, there are some major challenges around evaluations, overall reliability, and issues inherited from the language models powering each agent.

\subsection{Challenges with Agent Evaluation}
While LLMs are evaluated on a standard set of benchmarks designed to gauge their general understanding and reasoning capabilities, the benchmarks for agent evaluation vary greatly.

Many research teams introduce their own unique agent benchmarks alongside their agent implementation which makes comparing multiple agent implementations on the same benchmark challenging. Additionally, many of these new agent-specific benchmarks include a hand-crafted, highly complex, evaluation set where the results are manually scored \cite{chen_agentverse_2023}. This can provide a high-quality assessment of a method's capabilities, but it also lacks the robustness of a larger dataset and risks introducing bias into the evaluation, since the ones developing the method are also the ones writing and scoring the results. Agents can also have problems generating a consistent answer over multiple iterations, due to variability in the models, environment, or problem state. This added randomness poses a much larger problem to smaller, complex evaluation sets.

\subsection{Impact of Data Contamination and Static Benchmarks}
Some researchers evaluate their agent implementations on the typical LLM benchmarks. Emerging research indicates that there is significant data contamination in the model's training data, supported by the observation that a model's performance significantly worsens when benchmark questions are modified \cite{golchin_time_2024,zhu_dyval_2024,zhu_dyval2_2024}. This raises doubts on the authenticity of benchmark scores for both the language models and language model powered agents.

Furthermore, researchers have found that ``As LLMs progress at a rapid pace, existing datasets usually fail to match the models' ever-evolving capabilities, because the complexity level of existing benchmarks is usually static and fixed'' \cite{zhu_dyval2_2024}. To address this, work has been done to create dynamic benchmarks that are resistant to simple memorization  \cite{zhu_dyval_2024,zhu_dyval2_2024}. Researchers have also explored the idea of generating an entirely synthetic benchmark based on a user's specific environment or use case \cite{lei_s3eval_2023,wang_benchmark_2024}. While these techniques can help with contamination, decreasing the level of human involvement can pose additional risks regarding correctness and the ability to solve problems.

\subsection{Benchmark Scope and Transferability}
Many language model benchmarks are designed to be solved in a single iteration, with no tool calls, such as MMLU or GSM8K \cite{cobbe_training_2021,hendrycks_measuring_2021}. While these are important for measuring the abilities of base language models, they are not good proxies for agent capabilities because they do not account for agent systems' ability to reason over multiple steps or access outside information. StrategyQA improves upon this by assessing models' reasoning abilities over multiple steps, but the answers are limited to Yes/No responses \cite{geva_did_2021}. As the industry continues to pivot towards agent focused use-cases additional measures will be needed to better assess the performance and generalizability of agents to tasks involving tools that extend beyond their training data.

Some agent specific benchmarks like AgentBench evaluate language model-based agents in a variety of different environments such as web browsing, command-line interfaces, and video games \cite{liu_agentbench_2023}. This provides a better indication for how well agents can generalize to new environments, by reasoning, planning, and calling tools to achieve a given task. Benchmarks like AgentBench and SmartPlay introduce objective evaluation metrics designed to evaluate the implementation's success rate, output similarity to human responses, and overall efficiency \cite{liu_agentbench_2023,wu_smartplay_2024}. While these objective metrics are important to understanding the overall reliability and accuracy of the implementation, it is also important to consider more nuanced or subjective measures of performance. Metrics such as efficiency of tool use, reliability, and robustness of planning are nearly as important as success rate but are much more difficult to measure. Many of these metrics require evaluation by a human expert, which can be costly and time consuming compared to LLM-as-judge evaluations.

\subsection{Real-world Applicability}
Many of the existing benchmarks focus on the ability of Agent systems to reason over logic puzzles or video games \cite{liu_agentbench_2023}. While evaluating performance on these types of tasks can help get a sense of the reasoning capabilities of agent systems, it is unclear whether performance on these benchmarks translates to real-world performance. Specifically, real-world data can be noisy and cover a much wider breadth of topics that many common benchmarks lack.

One popular benchmark that uses real-world data is WildBench, which is sourced from the WildChat dataset of 570,000 real conversations with ChatGPT \cite{zhao2024inthewildchat}. Because of this, it covers a huge breadth of tasks and prompts. While WildBench covers a wide range of topics, most other real-world benchmarks focus on a specific task. For example, SWE-bench is a benchmark that uses a set of real-world issues raised on GitHub for software engineering tasks in Python \cite{jimenez_swe-bench_2023}. This can be very helpful when evaluating agents designed to write Python code and provides a sense for how well agents can reason about code related problems; however, it is less informative when trying to understand agent capabilities involving other programming languages.

\subsection{Bias and Fairness in Agent Systems}
Language Models have been known to exhibit bias both in terms of evaluation as well as in social or fairness terms \cite{gallegos_bias_2024}. Moreover, agents have specifically been shown to be ``less robust, prone to more harmful behaviors, and capable of generating stealthier content than LLMs, highlighting significant safety challenges'' \cite{tian_evil_2024}. Other research has found ``a tendency for LLM agents to conform to the model's inherent social biases despite being directed to debate from certain political perspectives'' \cite{taubenfeld_systematic_2024}. This tendency can lead to faulty reasoning  in any agent-based implementation.

As the complexity of tasks and agent involvement increases, more research is needed to identify and address biases within these systems. This poses a very large challenge to researchers, since scalable and novel benchmarks often involve some level of LLM involvement during creation. However, a truly robust benchmark for evaluating bias in LLM-based agents must include human evaluation.

\section{Conclusion and Future Directions}

The AI agent implementations explored in this survey demonstrate the rapid enhancement in language model powered reasoning, planning, and tool calling. Single and multi-agent patterns both show the ability to tackle complex multi-step problems that require advanced problem-solving skills. The key insights discussed in this paper suggest that the best agent architecture varies based on use case. Regardless of the architecture selected, the best performing agent systems tend to incorporate at least one of the following approaches: well defined system prompts, clear leadership and task division, dedicated reasoning / planning- execution - evaluation phases, dynamic team structures, human or agentic feedback, and intelligent message filtering. Architectures that leverage these techniques are more effective across a variety of benchmarks and problem types.

While the current state of AI-driven agents is promising, there are notable limitations and areas for future improvement. Challenges around comprehensive agent benchmarks, real world applicability, and the mitigation of harmful language model biases will need to be addressed in the near-term to enable reliable agents. By examining the progression from static language models to more dynamic, autonomous agents, this survey aims to provide a holistic understanding of the current AI agent landscape and offer insight for those building with existing agent architectures or developing custom agent architectures.

\printbibliography

\end{document}